# MENZERATH–ALTMANN LAW
# FOR SYNTACTIC STRUCTURES IN UKRAINIAN

Solomija Buk[1)] and Andrij Rovenchak[2)]

**Abstract:**

*In the paper, the definition of clause suitable for an automated processing of a Ukrainian text is proposed. The Menzerath–Altmann law is verified on the sentence level and the parameters for the dependences of the clause length counted in words and syllables on the sentence length counted in clauses are calculated for* Perekhresni Stežky (The Cross-Paths)*, a novel by Ivan Franko.*

## 1. INTRODUCTION

In 1928, Paul Menzerath observed the relation between word and syllable lengths: the average syllable length decreased as the number of syllables in the word grew. In the general form, such a dependence can be formulated as follows: the longer is the construct the shorter are its constituents. Later on, this fact was put in a mathematical form by Gabriel Altmann [1]. Now it is known as the Menzerath–Altmann law and is considered to be one of the general linguistic laws with evidences reaching far beyond the linguistic domain itself [2].

The mentioned relationship is studied on various levels of language units, such as syllable–word, morpheme–word, etc. While the word–sentence seems to be the most straightforward generalization on the syntactic level, it appears that in fact an intermediate unit must be introduced in this scheme [3, p. 283]. Usually, this intermediate unit are thought to be phrases or clauses, which are direct constituents of the sentence [4]. We would like to note, however, that the notion of clause is not well elaborated in Eastern European linguistic traditions [5], including Ukrainian (cf. syntax-related entries in [6]).

In this paper, we focus on the study of the novel *Perekhresni Stežky* (*The Cross-Paths*) by the Ukrainian writer Ivan Franko, for which the validity of Menzerath–Altmann law was previously confirmed on the syllable–word level [7]. In the next section, the definitions will

---

[1)] Dr. S. N. Buk, Department for General Linguistics, Ivan Franko National University of Lviv, 1 Universytetska St., Lviv, UA-79000, Ukraine, phone +380 32 2964756, e-mail: solomija@gmail.com.
[2)] Dr. A. A. Rovenchak, Department for Theoretical Physics, Ivan Franko National University of Lviv, 12 Drahomanov St., Lviv, UA-79005, Ukraine, phone +380 32 2550443, e-mail: andrij@ktf.franko.lviv.ua, andrij.rovenchak@gmail.com.

be given, including the clause definition suitable for the automatic text processing. Section 3 contains the results of numerical analysis. The conclusions are given in Section 4.

## 2. DEFINITIONS AND CLAUSE COUNT

In the paper, the following definitions are used:

- **Word** is a letter or alphanumeric sequence between two spaces (or punctuation marks)
- **Sentence** is a sequence of words between two delimiters (".", "!", "?", "…" *if followed by a capital letter of the next word*) (cf. [8]).

Several definitions for **clause** are available in the literature:

- word sequence with subject and predicate (like a sentence and unlike a syntagma) but which is contained in a sentence [9, p. 173];
- a linguistic structure that designates this kind of conceptually autonomous process, created through the elaboration of the participants in a temporal relation [10, p. 413];
- a unit of grammatical organization smaller than the sentence, but larger than phrases, words or morphemes [11, p. 74];
- a part of a sentence having a subject and predicate of its own [12, p. 183];
- a syntactic unit consisting of subject and predicate which alone forms a **simple sentence** and in combination with others forms a **compound sentence** or **complex sentence** [13, Vol. 10, p. 1501].

The notion of clause is traditionally connected with the presence of finite verb [14, s. 162]. Such an approach, however, fails in the case of Ukrainian due to several reasons explained below. Note that recently the violation of Menzerath–Altmann law was claimed by Maria Roukk for Russian having similar syntactic structure [15].

As there is no formal definition of clause suitable for Ukrainian we propose to define the number of clauses in a Ukrainian text by the following scheme counting for each sentence the number of:

- all verbal forms except an Infinitive (including gerund; see next for participles) — $N_1$;
- participles preceded by a comma (i.e., only those being in a post-position) — $N_2$;
- the so called 'predicative words' (*присудкові слова* in Ukrainian): **треба, нема, можна, годі, жаль, слід, шкода, непереливки, нічого, нічóго** — $N_3$;
- dashes (*tiret*) standing for the missing verb in compound predicates — $N_4$;
- conjunctions preceded by a comma — $N_C$.

$$\text{Number of clauses} = \max(N_1 + N_2 + N_3 + N_4;\ N_C + 1) \qquad (1)$$

According to this definition, the minimal number of clauses in a sentence is one. When a conjunction preceded by a comma appears, it usually means the presence of an additional predicative center and thus increasing the number of clauses in a compound sentence to at least two.

An expression containing a participle in a post-position can be normally converted into a subordinate clause. In other case, such a participle plays the role of an attribute. The following sentence can be an illustration: *Він стояв на тротуарі всміхнений, спотілий, з капелюхом, зсуненим на потилицю,…(He stood on the pavement smiling, perspiring, with the hat shifted on the back of his head…)*. In this example, the wavy underlining corresponds to an attribute, while the double line underlines the 'clause' participle. Note that the word *спотілий* is also preceded by a comma and can thus lead to the overestimation of the clause number in the automated processing.

A special attention paid to dashes is caused by a specific feature of Ukrainian grammar allowing for the omission of the verb *бути* (*to be*) in Present form [6, p. 634]. For instance, the sentence '*Зате совітник М. — картяр.*' translated as '*But councilor M. is a gambler*' literally reads '*But councilor M. — gambler*'. Some manual preprocessing must be done to distinguish such dashes.

## 3. RESULTS OF NUMERICAL ANALYSIS

We have made the analysis of *Perekhresni Stežky* (*The Cross-Paths*), a novel by Ivan Franko. This text was previously studied in detail, its frequency dictionary was compiled [16] and the online concordance is now available [17].

The text of the novel was tagged for parts of speech (PoS) allowing thus for the application of the scheme presented in previous section.

We have performed a manual count of clauses for Chapter I (of the total of 60) of the novel. The comparison with an automatic clause count is given in Table 1 below.

Table 1: Comparison of manual and automatic techniques for the clause count.

| Clauses per sentence | Clause length (manual) | Clause length (automatic) |
|---|---|---|
| 1 | 4.38 | 4.42 |
| 2 | 3.94 | 4.52 |
| 3 | 4.79 | 5.31 |
| 4 | 5.50 | 5.88 |
| 5 | 0.00 | 0.00 |
| 6 | 5.33 | 5.33 |
| 7 | 6.86 | 6.86 |
| 8 | 4.41 | 4.41 |

It appears that in Chapter I for 22 sentences of 116 (19%) the number of clauses was counted a bit incorrectly, almost all of them found in the direct speech. The reason is seen in the fact that the direct speech contains many incomplete sentences, in which no clause – according to our scheme – is marked explicitly.

We have made the analysis of the dependence of clause length (measured both in words and in syllables) on the sentence length measured in clauses. The results are summarized in Table 2.

It can be noticed that the clause length $f(x)$ counted in words is related to the sentence length $x$ counted in clauses by such a formula [18]:

$$f(x) = Ax^b e^{-c/x} \qquad (2)$$

with the following values of the fitting parameters:

$$A = 6.80 \pm 0.42$$
$$b = -0.11 \pm 0.03$$
$$c = 0.32 \pm 0.07$$
$$\chi^2 / N = 0.0066$$

It is known that expression (2) represents on of the generalizations of Menzerath–Altmann law [19].

Table 2: Sentence–clause data.

| Clauses per sentence | Clause length in words | | Clause length in syllables | | Number of sentences | |
|---|---|---|---|---|---|---|
| | observed | estimated | observed | estimated | observed | estimated |
| 1 | 4.96 | 4.94 | 9.82 | 9.81 | 3930 | 3929 |
| 2 | 5.34 | 5.39 | 10.97 | 11.04 | 2190 | 2201 |
| 3 | 5.47 | 5.44 | 11.45 | 11.32 | 1168 | 1150 |
| 4 | 5.36 | 4.42 | 11.30 | 11.38 | 597 | 586.4 |
| 5 | 5.45 | 5.38 | 11.29 | 11.37 | 259 | 295.3 |
| 6 | 5.41 | 5.34 | 11.45 | 11.34 | 142 | 147.6 |
| 7 | 5.34 | 5.29 | 11.50 | 11.29 | 82 | 73.4 |
| 8 | 5.13 | 5.25 | 10.87 | 11.23 | 44 | 36.4 |
| 9 | 5.32 | 5.20 | 11.31 | 11.18 | 22 | 18.0 |
| 10 | 5.78 | 5.17 | 12.10 | 11.12 | 9 | 8.9 |
| 11 | 5.58 | 5.13 | 12.42 | 11.07 | 6 | 4.4 |
| 12 | 4.17 | 5.09 | 8.25 | 11.02 | 1 | 2.2 |
| 13 | 8.42 | 5.06 | 19.50 | 10.97 | 2 | 1.1 |
| 14 | 4.57 | 5.03 | 9.50 | 10.93 | 1 | 0.52 |
| 15 | 0.00 | 5.00 | 0.00 | 10.88 | 0 | 0.25 |
| 16 | 6.09 | 4.97 | 13.81 | 10.84 | 2 | 0.12 |

It is possible to fit with the same functions also the sentence–syllable relations. Such results may be useful for comparisons between various languages, in particular with those having no apparent word-division. For instance, in languages using the Devanagari script the so-called breath groups are marked in writing and there are no word-boundaries [20, p. 126], special marks are used to separate clauses and sentences (but not words) in Tibetan [20, p. 503], etc. The values of the fitting parameters for the sentence–syllable case are listed below:

$$A = 13.9 \pm 0.9$$
$$b = -0.08 \pm 0.03$$
$$c = 0.35 \pm 0.08$$
$$\chi^2 / N = 0.042$$

The results of fitting are represented in Figs 1,2.

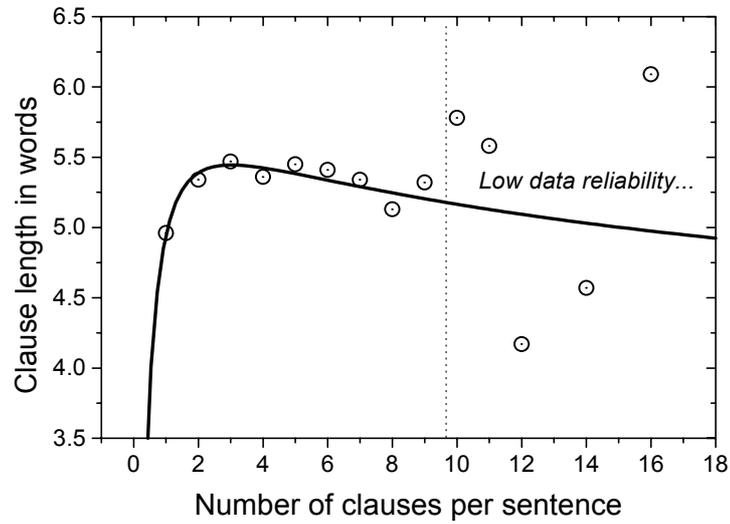

Fig. 1. The clause length in words versus the number of clauses per sentence.

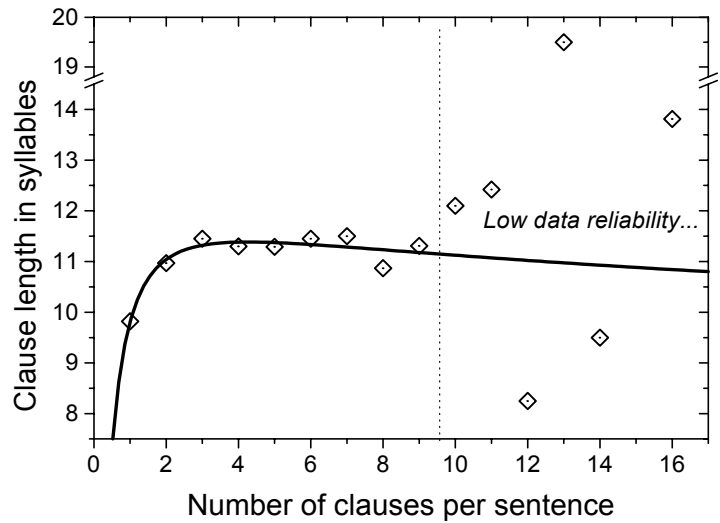

Fig. 2. The clause length in syllables versus the number of clauses per sentence.

While searching for a simple model for the dependence of the number of sentences $g(x)$ versus the number of constituting clauses $x$ we found that the negative binomial distribution is an appropriate one:

$$g(x) = \binom{r+x-2}{x-1} p^r (1-p)^{x-1}. \qquad (3)$$

The following are the values of the fitting parameters (see Fig. 3 for the graphical comparison):

$$p = 0.515 \pm 0.006$$
$$r = 1.16 \pm 0.02$$
$$\chi^2 / N = 2 \cdot 10^{-6}$$

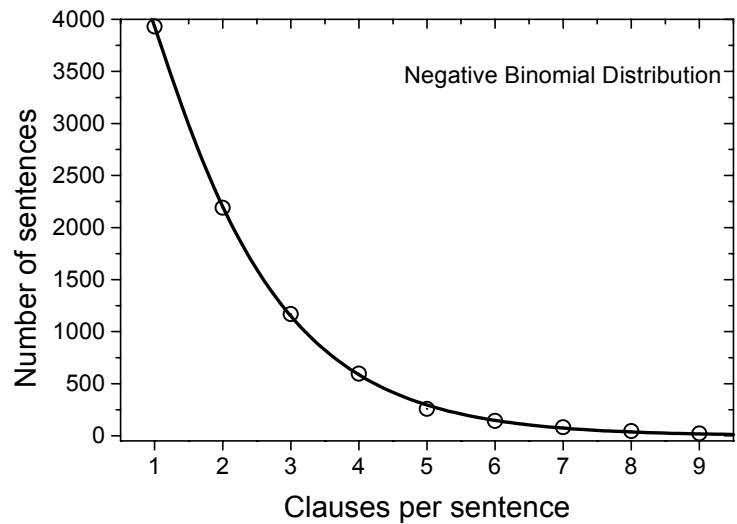

Fig. 3. The number of sentences versus the number of clauses per sentence.

## 4. CONCLUDING REMARKS

From the consideration given in this paper several conclusions can be made. First of all, it is clear that the specification of the 'clause' notion is needed, especially in the direct speech. The proposed scheme for the clause count highly relies on the PoS-tagged material and this slows its implementation. Some improvements must be introduced into the scheme to avoid over- and underestimation of the number of clauses. A special treatment for a correct accounting of the compound predicates lacking the verbal component must be elaborated. One should analyze more texts to confirm the models. Also, the dependence of fitting parameters on style / authorship must be studied.

## ACKNOWLEDGEMENTS

We highly appreciate e-discussions and useful suggestions made by *Peter Grzybek* on the subject of the presented work in its various aspects.